\pdfoutput=1

\documentclass[11pt]{article}

\usepackage{emnlp2021}

\usepackage{times}
\usepackage{latexsym}
\usepackage{xspace}

\usepackage[T1]{fontenc}
\usepackage[utf8]{inputenc}

\usepackage{microtype}

\usepackage{amsmath}
\usepackage{algorithm}
\usepackage[noend]{algpseudocode}
\usepackage{dsfont}
\usepackage{booktabs}
\usepackage{multirow}
\usepackage{siunitx}
\usepackage{xcolor}
\newcommand\indicator{\mathds{1}}

\usepackage{cleveref}
\crefname{section}{\S}{\S\S}
\Crefname{section}{\S}{\S\S}
\crefformat{section}{\S#2#1#3}
\crefname{algorithm}{Alg}{}
\crefname{algorithm}{Alg}{}
\crefname{line}{Line}{}
\crefname{equation}{Eq.}{}
\Crefname{equation}{}{}
\crefname{table}{Table}{}
\Crefname{table}{Table}{}
\crefname{plt}{Figure}{}

\usepackage[disable]{todonotes}
\makeatletter
\newcommand*\iftodonotes{\if@todonotes@disabled\expandafter\@secondoftwo\else\expandafter\@firstoftwo\fi}  %
\makeatother

\newcommand{\note}[4][]{\todo[author=#2,color=#3,size=\scriptsize,fancyline,caption={},#1]{#4}} %
\newcommand{\ryan}[2][]{\note[#1]{ryan}{violet!40}{#2}}
\newcommand{\clara}[2][]{\note[#1]{clara}{orange}{#2}}
\newcommand{\liam}[2][]{\note[#1]{liam}{yellow}{#2}}
\newcommand{\response}[1]{\vspace{3pt}\hrule\vspace{3pt}\textbf{#1:}\xspace}

\newcommand{\xx}{\mathbf{x}}
\newcommand{\yy}{\mathbf{y}}
\newcommand{\calY}{\mathcal{Y}}

\newcommand{\vocab}{\mathcal{V}}
\newcommand{\vocabeos}{\overline\vocab}
\DeclareMathOperator*{\argmax}{argmax}
\newcommand{\eos}{\textsc{eos}\xspace}
\newcommand{\bos}{\textsc{bos}\xspace}
\newcommand{\score}{\mathrm{score}}
\newcommand{\xsum}{\textsc{Xsum}\xspace}
\newcommand{\rouge}{\textsc{rouge}\xspace}
\newcommand{\bertscore}{\textsc{berts}core\xspace}
\newcommand{\berts}{\textsc{berts}\xspace}
\newcommand{\factscore}{\textsc{facts}core\xspace}
\newcommand{\factcc}{FactCC\xspace}
\newcommand{\ent}{\mathrm{H}}

\newcommand{\cnn}{\textsc{CNN/DM}\xspace}
\newcommand{\ourdecoding}{\textsc{CPMI}\xspace}
\newcommand{\bartmodel}{\textsc{bartS2S}\xspace}
\newcommand{\transmodel}{\textsc{tranS2S}\xspace}
\newcommand{\mathcheck}[1]{{\color{black} #1}}
\newcommand{\defeq}[0]{\mathrel{\stackrel{\textnormal{\tiny def}}{=}}}

\newcommand*{\Scale}[2][4]{\scalebox{#1}{$#2$}}
\newcommand{\smallpm}[1]{\Scale[0.8]{\pm\color{blue}{#1}}}

\makeatletter
\newcommand{\thickhline}{%
    \noalign {\ifnum 0=`}\fi \hrule height 1pt
    \futurelet \reserved@a \@xhline
}
\makeatother

\usepackage{inconsolata}  
\usepackage{amssymb,graphicx}
\usepackage{tikz}
\usepackage{pgfplots}
\usepackage{color}
\usepackage{xcolor}	
\everypar{\looseness=-1}

\title{Mutual Information Alleviates Hallucinations in Abstractive Summarization}

\author{Liam van der Poel~\;~Ryan Cotterell~\;~Clara Meister \\
\fcolorbox{white}{white}{
\texttt{\href{mailto:lvander@ethz.ch}{lvander@ethz.ch}} ~\;~$\{$\texttt{\href{mailto:ryan.cotterell.inf.ethz.ch}{ryan.cotterell}, }\texttt{\href{mailto:clara.meister@inf.ethz.ch}{ clara.meister}}$\}$\texttt{@inf.ethz.ch}
} \\ 
\setlength{\fboxsep}{2.5pt}
\setlength{\fboxrule}{2.5pt}
\fcolorbox{white}{white}{
\includegraphics[width=.15\linewidth]{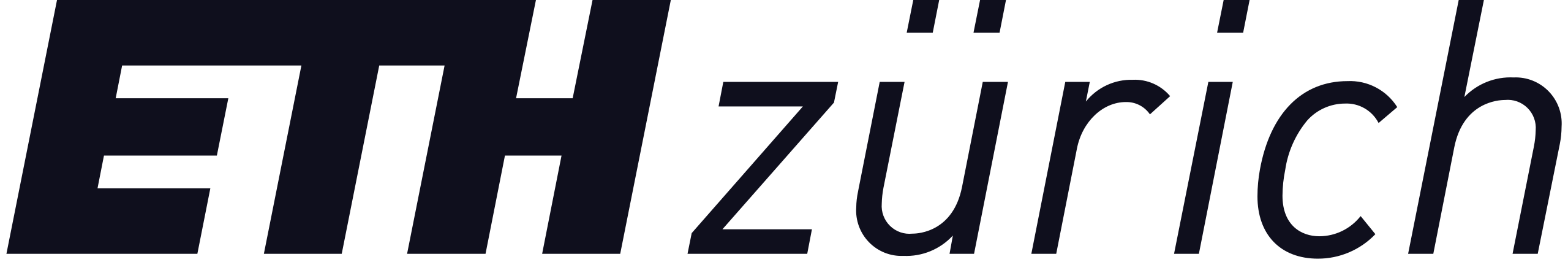}
}
}

\begin{document}
\maketitle
\begin{abstract}
Despite significant progress in the quality of language generated from abstractive summarization models, these models still exhibit the tendency to hallucinate, i.e., output content not supported by the source document. A number of works have tried to fix---or at least uncover the source of---the problem with limited success. In this paper, we identify a simple criterion under which models are significantly more likely to assign more probability to hallucinated content during generation: high model uncertainty. This finding offers a potential explanation for hallucinations: models default to favoring text with high marginal probability, i.e., high-frequency occurrences in the training set,  when uncertain about a continuation. It also motivates possible routes for real-time intervention during decoding to prevent such hallucinations. We propose a decoding strategy that switches to optimizing for pointwise mutual information of the source and target token---rather than purely the probability of the target token---when the model exhibits uncertainty. Experiments on the \xsum dataset show that our method decreases the probability of hallucinated tokens while maintaining the \rouge and \berts scores of top-performing decoding strategies.

\vspace{1.5em}
\hspace{.5em}\includegraphics[width=1.25em,height=1.25em]{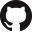}\hspace{.75em}\parbox{\dimexpr\linewidth-2\fboxsep-2\fboxrule}{\url{https://github.com/VanderpoelLiam/CPMI}}
\vspace{-.5em}
\end{abstract}

\section{Introduction}
Abstractive summarization, the task of condensing long documents into short summaries, has a number of applications, such as providing overviews of news articles or highlighting main points in technical documents. Abstractive summarization is usually performed using probabilistic text generators \cite{goyal2020factuality, mao2020constrained, kryscinski2019evaluating}, which have shown a strong ability to produce fluent, human-like text \cite{ baevski2019adaptive, radford2019language, brown2020language}. However, these models have been observed to \textbf{hallucinate} facts, i.e., add information to the output that was not present in the original text. This behavior is problematic, as presenting users with unsubstantiated content can lead to undesirable effects, such as the spread of misinformation \cite{bender2021dangers, abid2021large, liang2021towards}. Some works have attributed this phenomenon to the specific training corpora for these models, in which ground-truth summaries often contain outside information that may not have been directly deducible from the original text \cite{maynez2020faithfulness,zhou2021detecting}. Others have pointed to model architectures or training strategies \cite{voita2021analyzing, wang2020exposure, kang2020improved}. While these works have given us an improved understanding of the cause of hallucinations, there still does not exist an efficient and robust set of techniques for identifying and preventing them during the generation process. 

This work aims to first provide a simple criterion indicating when a model is more likely to assign higher probability to content not necessarily derived from the source document.  
Specifically, we link the start of a hallucination during generation to high model uncertainty about the next token, which we quantify by conditional entropy. 
We hypothesize that hallucinations may be due to a tendency of models to default to placing probability mass on tokens that appeared frequently in the training corpus, a  behavior by language models  previously observed in several natural language processing (NLP) tasks \cite{kobayashi-etal-2020-attention,wei2021frequency}. As a consequence, generations with hallucinations would still be viable candidates, as standard decoding strategies for summarization optimize purely for the probability of the generation. 
We propose an alternative decoding strategy to combat this behavior: When a model exhibits high uncertainty, we  change our decoding objective to pointwise mutual information between the source document and target token \cite[PMI;][]{li2016diversitypromoting,takayama2019relevant}, encouraging the model to prioritize tokens relevant to the source document. While changing completely to the PMI objective causes a drop of $3.13\%$ in \rouge-L scores, this conditional and temporary change leads to  only a $0.977\%$ drop in \rouge-L while increasing factuality according to the \factscore metric.  

\ryan{The thing that is running through my head now is why not use PMI the entire time? What is the advantage of switching?\response{liam}{added explanation with data}}

In experiments, we first observe a strong correlation between conditional entropy and the start of a hallucination on an annotated subset of the \xsum dataset \cite{maynez2020faithfulness}. 
We next score the targets in the annotated subset under both the standard log-probability objective and \ourdecoding, and observe that the revised log-probability of hallucinated tokens under the \ourdecoding objective is indeed lower \liam{Changed PMI to \ourdecoding here, I think this is a typo}. 
Finally, we find that our proposed decoding strategy maintains \rouge and \berts scores.\ryan{Did you run just a PMI baseline?\response{liam}{yes, added to previous paragraph} \response{ryan} where is it? Can you put it in the table.\response{liam} This was only run as an preliminary experiment for \transmodel on the full validation set, I do not have the equivalent data for the table}

\section{Preliminaries}
In this work, we consider probabilistic models for abstractive summarization. Explicitly, we consider models with distribution $p(\yy \!\mid\! \xx)$, where $\xx$ is the source document that we wish to summarize and $\yy = \langle y_0, \dots, y_T\rangle$\ryan{I played around with the indexing since it's weird to me to make $\bos$ part of the string. This a note to make sure it's all coherent.\response{clara}{I changed this back. It felt too confusing with the definition of $\yy_{<t}$. Happy to find a compromise though!}} is a string, represented as a sequence of tokens from the model's vocabulary $\vocab$. The set of  valid sequences $\calY$ is then defined as all sequences $\yy$ such that $y_0 \defeq \bos$ and $y_T \defeq \eos$, the beginning- and end-of-sequence tokens, respectively, and $y_t \in \vocab$ for $0 < t <T$.
Note that standard models are locally normalized, i.e., they provide a probability distribution over $\vocabeos \defeq \vocab \cup \{\eos\}$ at time step $t$ given the source document and prior context $p(\cdot \mid \yy_{<t}, \xx)$.  The probability of an entire string $\yy$ can then be computed as $p(\yy \mid \xx) = \prod_{t=1}^T p(y_t \mid \yy_{<t}, \xx)$, where for shorthand we define $\yy_{<t} \defeq \langle y_0, \dots, y_{t-1}\rangle$. 

Generation from $p$ is performed token-by-token due to the autoregressive natures of most language generators. We typically seek to generate a string that maximizes some score function
\begin{equation}
    \mathcheck{\yy^\star = \argmax_{\yy \in \calY} \,\, \score(\yy \mid \xx)}
\label{eq:opt-problem}
\end{equation}
In the case of probabilistic models, this function is often simply $\score(\yy \mid \xx) = \log p(\yy \mid \xx)$, i.e., we want to generate a high probability string $\yy$. Note that searching over the entire space $\calY$ is usually infeasible (or at least impractical) due to the non-Markovian nature of most neural models. Thus we often use an approximate search algorithm such as beam search, as given in \cref{alg:beam}, that optimizes for our score function somewhat greedily. This procedure meshes well with the use of $\log p$ as the score function since it can be decomposed as the sum of individual token log-probabilities, i.e., we can instead consider a token-wise score function $\mathcheck{\score(y \mid \yy_{<t}, \xx) = \log p(y \mid \yy_{<t}, \xx)}$ using the fact that $\score(\yy\!\mid\!\xx) = \sum_{t=1}^T\score(y \!\mid\! \yy_{<t}, \xx)$. We only consider decoding strategies for score functions that can be decomposed in this manner.\looseness=-1

\algrenewcommand{\algorithmiccomment}[1]{\hskip1em$\rightarrow$ \footnotesize#1 \normalsize}
\begin{algorithm}[!tb]
\textbf{Input:} $\xx$: source document\\
\hspace*{2.7em} $k$: maximum beam size \\
\hspace*{2.7em} $n_{\mathrm{max}}$: maximum hypothesis length \\
\hspace*{2.7em} $\score(\cdot\mid \cdot)$: scoring function
\begin{algorithmic}[1]
\State $B_0 \gets \{ \langle 0, \bos \rangle \}$ \Comment{{\color{gray}\emph{beam set}}}
\For{ $t \in \{ 1, \dots, n_{\mathrm{max}} \}$ }
    \State $B \gets \emptyset$ 
    \For{$ \langle s, \yy \rangle \in B_{t-1}$}
    \For{$y \in \vocabeos$}
        \State $s \gets \score(\yy \circ y\mid \xx)$ \label{line:bs-score-eval}
        \State $B.\mathrm{add}( \langle s, \yy \circ y \rangle)$
    \EndFor
    \EndFor
    \State $B_t \gets B.\mathrm{top}(k)$
    \EndFor
\State \Return $B_{n_{\mathrm{max}}}.\mathrm{max}()$
\end{algorithmic}
\caption{Standard beam search.\footnotemark $\,\circ$ used to represent string concatenation.}
\label{alg:beam}
\end{algorithm}
\addtocounter{footnote}{-1} %
\stepcounter{footnote}\footnotetext{Pseudocode taken from \citet{meister-etal-2020-best}.}
\paragraph{Evaluation.} Abstractive summarization systems are usually evaluated using automatic metrics, such as \rouge \cite{lin2004rouge}. While \rouge generally correlates poorly with human judgments \cite{maynez2020faithfulness, fabbri2020summeval} and is only weakly correlated with factuality,\footnote{\rouge-2 on \xsum has $0.17$ Pearson and $0.14$ Spearman correlation \cite{deutsch2021,pagnoni2021understanding}} it is quick to compute, making it useful for quickly testing modeling choices. 
Recently, entailment metrics \citep[\factcc;][]{kryscinski2019evaluating} and contextual embedding methods \citep[\bertscore;][]{zhang2019bertscore} have surfaced as reasonable indicators of factuality.%

\section{Finding and Combating Hallucinations}
It is not well understood \emph{when} summarization models start to hallucinate, i.e., when they start to place high probability on continuations that are unfaithful (not entailed by the information presented in the source document). In this work, we hypothesize that such moments correlate with high model uncertainty. 
In other problem settings, it has been observed that NLP models default to placing an inappropriately large portion of probability mass on high-frequency (with respect to the training corpus) tokens; this is especially the case when making predictions for data points of a type that the model has not had much exposure to during training \cite{kobayashi-etal-2020-attention,wei2021frequency}. In this same setting, models often have high (epistemic) uncertainty about their predictions \cite{Hllermeier2021AleatoricAE}. We extrapolate on these findings and posit that summarization models may  score highly more marginally likely---but perhaps unrelated---tokens in settings for which they are not well-calibrated. \clara{In general, revisit this paragraph: need to link frequency to marginal likelihood. Basically say something like we can extrapolate that models do the same for frequently observed sequences of text, i.e., those with high marginal likelihood \response{liam} }

Fortunately, both model certainty and marginal likelihood have quantifications that can be easily computed at any given point in the decoding process, making it possible to test for relationships between these quantities and the start of \textbf{hallucinations}. Specifically, we can use the standard equation for Shannon entropy with our conditional distribution to quantify model uncertainty  at time step $t$: \mathcheck{$\ent(p(\cdot  \! \mid\! \yy_{<t}, \xx)) = -\sum_{y\in\vocabeos} p(y\! \mid\!  \yy_{<t}, \xx) \log p(y \mid  \yy_{<t}, \xx)$}. Entropy is not a holistic measure of uncertainty,\footnote{While we may also use techniques like MC dropout \cite{gal2016dropout} to quantify model uncertainty, such metrics would only capture epistemic uncertainty. Entropy on the other hand, should provide a quantification of both aleatoric and epistemic uncertainty.} but our use  of is motivated by previous research that has likewise employed it to quantify the uncertainty of model predictions in classification \cite{gal2016dropout} and summarization \cite{xu2020understanding} tasks.  %
Further, we can directly compute marginal probabilities $p(y \!\mid\! \yy_{<t})$ using a language model---this value  quantifies how likely a continuation $y$ is irrespective of the source.\mathcheck{\footnote{As \mathcheck{$p(y\!\mid\!\yy_{<t}) = \sum_{\xx} p(y\!\mid\!\yy_{<t}, \xx)$}, the language model probability is equivalent to marginalizing over all source documents $\xx$.}}   

\subsection{Pointwise Mutual Information Decoding}
Under the premise that models are placing disproportionate probability mass on marginally likely, (i.e., frequent) tokens, the standard log-probability decoding objective is prone to favor generic continuations regardless of the input.
In order to alleviate the problem of generic outputs from neural conversation models, \citet{li2016diversitypromoting} propose maximizing for mutual information during decoding, which effectively introduces a penalty term for such candidates. Formally, they propose using the following score function in the problem of \cref{eq:opt-problem}: 
\begin{equation}\label{eq:mmi_opt}
    \score(\yy\!\mid\!\xx) = \log \frac{p( \xx, \yy)}{p(\xx)p(\yy)}
\end{equation}
\noindent which is the (pairwise) mutual information between the source $\xx$ and the target $\yy$. 
Note that this would be equivalent to optimizing for $\score(\yy\!\mid\!\xx) = \log p(\yy \mid \xx) - \log p(\yy)$.\footnote{This follows as \mathcheck{$\log\frac{p(\xx,\yy)}{p(\xx)p(\yy)} = \log \frac{p(\yy \mid \xx)}{ p(\yy)} = \log p(\yy \mid \xx) - \log p(\yy)$} after applying Bayes rule.} While this score function likewise decomposes over tokens, for the same reasons as discussed earlier, solving for the exact minimizer is computationally intractable. Thus we must still resort to approximate search algorithms. In practice, one can iteratively optimize for \emph{pointwise} mutual information (PMI): $\mathcheck{\log p(y \mid \yy_{<t}, \xx) - \log p(y \mid \yy_{<t})}$.\footnote{Other practical considerations are the introduction of the hyperparameter $\lambda$ to control the strength of the influence of the language model $p(\yy)$.}

Our proposed decoding strategy, \textbf{conditional PMI decoding} (\ourdecoding), uses the conditional entropy at a given time step to indicate when the pointwise decoding objective should be changed. 
This process can be formalized as follows. For a given (token-by-token) decoding strategy,  we use the pointwise score function:
\mathcheck{
\begin{align}
   \score(&y \mid  \yy_{<t}, \xx) =  \log p(y \mid \yy_{<t}, \xx) \\
    &\,- \lambda \cdot \indicator\Scale[0.8]{\Big\{ \ent(p(\cdot \mid \yy_{<t}, \xx)) \geq \tau \Big\}} \cdot \log p(y \mid \yy_{<t}) \nonumber
\end{align}
}
\noindent In words, when $\ent(p(\cdot\! \mid\! \yy_{<t}, \xx))$ is above a certain threshold $\tau$, we subtract a term for the marginal log-probability of the token, i.e., we change from the standard token-wise log-probability objective to PMI. %

\section{Related Work}\label{sec:related}
\paragraph{Understanding hallucinations.} Several prior works have tried to identify the cause of hallucinations in various natural language generation tasks, along with methods for alleviating them. For example, both \citet{wang2020exposure} and \citet{voita2021analyzing} suggest that exposure bias, i.e., the failure of a model to predict accurate continuations following its own generations rather than the ground-truth context as a result of the discrepancy between procedures at training and inference, leads to hallucinations, as it causes the model to over-rely on target contributions when decoding. They propose using minimum risk training (MRT), which can alleviate exposure bias, to make models more robust. 
However these results show only a tentative connection to exposure bias, and are based on models for neural machine translation (NMT) rather than summarization. Other works have shown that pre-training and training on more data generate summaries more faithful to the source \cite{voita2021analyzing, maynez2020faithfulness}. In contrast to these works, our method does not require any changes to model training. Rather, it intervenes during generation without the need to retrain the base model. 
\paragraph{Detecting hallucinations.} Other efforts aim to identify hallucinations rather than their cause. Token- \cite{zhou2021detecting} and sentence-level \cite{kryscinski2019evaluating} hallucination detection, as well as textual entailment systems \cite{goyal2020factuality} allow hallucinations to be identified after the generation process. Some techniques even aim to correct the unfaithful span by, e.g., replacing it with text from the source \cite{chen2021improving}. However, these approaches are all post-hoc. Our approach intervenes during decoding, allowing real-time hallucination detection and prevention. 
\paragraph{Decoding to avoid hallucinations.} Perhaps most in-line with this work, some prior work has modified the decoding procedure to avoid unfaithful outputs. Keyword-based methods extract keyphrases from the source and require that they appear in the summary \cite{mao2020constrained}. 
The focus attention mechanism \cite{aralikatte2021focus} biases the decoder towards tokens that are similar to the source. 
While this is similar to our approach, we use mutual information to bias our decoding algorithm away from high probability---but not necessarily relevant---candidates. Another difference is that our method only runs when model uncertainty, as quantified by conditional entropy, is high, so we only bias generation when necessary. 
Lastly, our approach is purely abstractive and does not require resorting to extractive methods. \ryan{What does that mean?\response{clara}{its summarization jargon :p there's extractive summarization and abstractive summarization, which are two different tasks}}
\paragraph{Mutual information decoding.}
Mutual information-based decoding techniques have proven to be helpful in a number of settings. For example, in zero-shot settings  \cite{holtzman2021surface} or for promoting diversity or relevance in neural dialogue models \cite{li2016diversitypromoting,takayama2019relevant}. Our work is the first to use mutual information to increase the faithfulness of summaries in abstractive summarization. 

\section{Experiments}
\paragraph{Data.}
We use the extreme summarization (\xsum) dataset \cite{narayan2018xsum}, which is composed of 226,711 British Broadcasting Corporation (BBC) articles and their single-sentence summaries. We use the same train--valid--test splits as the authors. %
A subset of these articles (500) from the test set are annotated, i.e., reference spans are labeled as faithful or not to the source article \cite{maynez2020faithfulness}. 
We further process these labels to obtain token-level hallucination labels.
\paragraph{Models.}
We use the Fairseq framework \cite{ott2019fairseq} for all of our experiments. 
We evaluate several models: a transformer based summarization model (\transmodel) trained on the \xsum dataset with the standard maximum log-likelihood objective as well as the BART summarization \clara{@Liam which version of BART? \response{liam}{summarization? It is the version from the paper I cited}\response{clara}{Is the checkpoint from the fairseq library? If so, we should link to the readme}\response{liam}done} model (\bartmodel) fine-tuned on \xsum \cite{lewis2019bart}.\footnote{\url{https://github.com/facebookresearch/fairseq/tree/main/examples/bart}} Lastly, for our language model $p(\yy)$, we train a Transformer-based language model.\footnote{We use both \xsum articles and summaries when training our LM as we found the summaries alone did not constitute enough data to train a well-performing LM.}

\paragraph{Decoding.} 
 We  generate summaries using \ourdecoding and beam search, as well as score existing summaries under the \ourdecoding objective.  We do a hyperparameter search to select the two hyperparameters $\lambda / \tau$ (see \cref{sec:appendix-implementation} for details). For evaluations, we would ideally use token-level faithfulness labels. However, we have only 500 such human annotated reference summaries \cite{maynez2020faithfulness}. To obtain labels for the generated text, and the 10,832 other reference summaries, we turn to automatic factuality detection \cite{zhou2021detecting}. This allows us to evaluate on metrics specific to the token label e.g., average probability of hallucinated tokens. \liam{Added this section about PMI decoding} We investigate solely using the PMI objective by allowing $\tau=0$ during our hyperparameter search.

\paragraph{Evaluation.}
Lacking a good operationalization of a hallucination, we cannot directly measure the fraction of hallucinated tokens in the generated summaries. In line with previous work, we rely on automatic metrics and human evaluations to estimate the incidence of hallucinations \cite{nie2019simple,maynez2020faithfulness,zhou2021detecting}. \ourdecoding is evaluated using standard summarization performance metrics (\rouge and \bertscore), factuality metrics (\factcc and \factscore), and an estimation of hallucination incidence based on scoring reference summaries with associated human evaluated hallucination labels. 
The \factcc metric computes a token-level binary factuality score over a collection of source/summary pairs and returns the mean \cite{pagnoni2021understanding}. It uses the binary entailment classifier of the same name \cite{kryscinski2019evaluating}. Thus we can produce a similar entailment metric using the factuality labeling generated by \citet{zhou2021detecting}, which we denote \factscore. The uncertainty values of our results denote standard error.

\begin{table}
\centering
\resizebox{\linewidth}{!}{
\begin{tabular}{l| c c | c c }
\toprule
 & \multicolumn{2}{c|}{\transmodel} & \multicolumn{2}{c}{\bartmodel} \\
\midrule
Metrics & Beam Search & \ourdecoding & Beam Search & \ourdecoding\\
\midrule
\rouge-L & \textbf{0.252} & 0.249       & \textbf{0.372} & \textbf{0.372} \\
\berts P & \textbf{0.901} & 0.897       & \textbf{0.926} & \textbf{0.926} \\
\berts R & \textbf{0.886} & 0.885       & \textbf{0.917} & \textbf{0.917} \\
\berts F1 & \textbf{0.893} & 0.891       & \textbf{0.922} & \textbf{0.922} \\
\midrule
\factscore & 0.155 & \textbf{0.167}      & 0.126 & \textbf{0.128}  \\
\factcc    & \textbf{0.221} & 0.193      & \textbf{0.232} & 0.227  \\
\bottomrule
\end{tabular}
}
\caption{ Performance and factuality metrics for both models and decoding procedures}
\label{tab:factuality-results}
\end{table}

\begin{table}
\centering
\resizebox{\linewidth}{!}{
\begin{tabular}{l| c c | c c }
\toprule
\multirow{2}{*}{Token Label} & \multicolumn{2}{c}{\transmodel} & \multicolumn{2}{c}{\bartmodel} \\
&$\Delta \score$ & $\Delta $ rank &$\Delta \score$ & $\Delta $rank   \\
\midrule
Non-Hallucinated        & $-0.35 \smallpm{0.03}$ & $987 \smallpm{ 91}$ & $-0.07 \smallpm{ 0.01}$& $275 \smallpm{ 114}$\\
Hallucinated            & $-0.50 \smallpm{ 0.01}$& $1228 \smallpm{ 42}$& $-0.07 \smallpm{ 0.01}$& $395 \smallpm{ 45}$\\
\hspace{3mm}Initial    & $-0.55 \smallpm{ 0.04}$& $1272 \smallpm{ 126}$& $-0.13 \smallpm{ 0.03}$& $552 \smallpm{ 134}$\\
\hspace{3mm}Subsequent & $-0.50 \smallpm{ 0.01}$& $1221 \smallpm{ 44}$& $-0.07 \smallpm{ 0.01}$& $376 \smallpm{47}$\\
\bottomrule
\end{tabular}
}
\caption{Change in average token score and ranking by ground-truth hallucination label for \ourdecoding compared to beam search.}
\label{tab:change-hallucination-results}
\end{table}

\subsection{Preliminary Analysis}
Using the 500 example subset of \xsum with factuality annotations \cite{maynez2020faithfulness}, we are able to determine whether a token is hallucinated, and further whether it is the first in a sequence of hallucinated tokens. Our preliminary investigations found that on average, the conditional entropy under our summarization model is higher for first hallucinated tokens relative to non-hallucinated tokens ($4.197 \pm0.065$ vs $3.689 \pm 0.021$ for \transmodel and $3.115	 \pm 0.051$ vs $2.390 \pm 0.013$ for \bartmodel). This suggests that hallucinations could be connected with model uncertainty, and that the start of hallucinations could be identified  when conditional entropy is above a certain threshold, the model defaults to a likely but perhaps unfaithful token. 

\subsection{Results}

We now perform our generation and scoring analyses, as outlined above. 

\paragraph{How are performance and factuality metrics impacted by \ourdecoding?} From \cref{tab:factuality-results} we see that for the \bartmodel model there is very little change to performance metrics. For the \transmodel model, performance metrics are slightly worse under \ourdecoding, but the largest change is still within the margin of error. This suggests \ourdecoding does not negatively impact the quality of generated sentences. While \factscore increases under \ourdecoding for both models, \factcc decreases. However, \factscore employs a model trained specifically for \xsum while \factcc employs a model trained only on \cnn. Consequently, we take the results of \factscore as a better indicator of the effect of our approach on hallucination incidence during generation. Examples of summaries generated with and without \ourdecoding are given in \cref{figure:xsum-example}.

\paragraph{What happens to known unfaithful tokens when scored under \ourdecoding?} \Cref{tab:change-hallucination-results} \clara{@Liam what are the $\pm$ values here? \response{liam}{added line at the end of the evaluation section}} shows how token-level score and ranking (where the highest-probability token is rank $1$ and lowest probability token is rank $|\vocab|$) change when \ourdecoding is used instead of the standard log-probability scoring function for 500 ground-truth summaries with human evaluated factuality labels. 
Overall, we see that for hallucinated tokens, the scores decrease and ranking increases, which is the desired behavior. 
This is particularly true for tokens at the start of an unfaithful span (denoted as Initial), for which we see a more significant impact on both models. 
E.g., for \bartmodel, the score decreases more for initial vs. non-hallucinated ($-0.13\pm0.03$ vs $-0.07\pm0.01$) and likewise rankings increasing more ($275 \pm114$ vs. $552\pm134$). While there is also an impact on non-hallucinated tokens, it is much less significant. For an appropriate choice of threshold though, it is likely that PMI will not be in action at times when non-hallucinated tokens would have been chosen, meaning this change should not be a concern. 

\section{Conclusion}
In this work, we link the start of a hallucination in abstractive summarization during generation to model uncertainty, as quantified by high conditional entropy, about next-token predictions. 
We then propose a decoding procedure, \ourdecoding, which switches the decoding objective to pointwise mutual information when model uncertainty is high to prevent hallucinations. Our method reduces the likelihood of generating unfaithful tokens while still outputting high-quality summaries. 
In the future it would be interesting to combine \ourdecoding decoding with post-hoc correction methods and other decoding procedures, to investigate if we can complement existing techniques mentioned in \cref{sec:related}.\looseness=-1
\section*{Limitations}
A clear limitation of this work is that the results have been shown only for English on the \xsum dataset, as this is the only open dataset with the annotations required for our set of experiments. Further work should consider other model architectures, and other datasets such as \cnn \cite{hermann2015cnn}. Further, we do not conduct human evaluations. Using human judges to obtain a qualitative assessment of the effect of \ourdecoding could provide additional data about the efficacy of the decoding procedure. However, we note that human judgment of the faithfulness of summaries is far from perfect \cite{clark2021gold}. %

There are issues with the \xsum dataset that may be confounders for results: some articles/summaries are in Gaelic, and previous work has shown that reference summaries often contain spans not directly inferable from the source article \cite{maynez2020faithfulness}. Limitations of the models themselves are that we truncate the source to 4096 tokens, so we lose information due to this training constraint.

\section*{Ethical Concerns} We do not foresee any ethical concerns with this work beyond those already documented in abstractive summarization systems and other text generators \citep{smiley-etal-2017-say,NEURIPS2019_3e9f0fc9,kreps2022all}.\looseness=-1

\bibliography{anthology,custom}
\bibliographystyle{acl_natbib}

\appendix
\clearpage
\section{Additional Results}
\label{sec:appendix-data}
\cref{tab:hallucination-results} contains the results of evaluating beam search and \ourdecoding on ground truth token labels that was processed to generate \cref{tab:change-hallucination-results}. \cref{tab:entropy-results} contains the full results of our preliminary analysis to correlate average conditional entropy of the summarization model by token labels. \cref{alg:beam} provides the standard beam search algorithm.

\begin{table*}
\centering
\resizebox{\linewidth}{!}{
\begin{tabular}{l| c c c c | c c c c }
\toprule
 & \multicolumn{4}{c|}{\transmodel} & \multicolumn{4}{c}{\bartmodel} \\
\midrule
\multirow{2}{*}{Token label} & \multicolumn{2}{c}{Beam Search} & \multicolumn{2}{c|}{\ourdecoding} & \multicolumn{2}{c}{Beam Search} & \multicolumn{2}{c}{\ourdecoding} \\
&$\score$ & rank &$\score$ & rank &$\score$ & rank &$\score$ & rank  \\
\midrule
Non-Hallucinated        & $-3.58 \pm 0.02$ & $8020 \pm 62$ & $-3.93 \pm 0.02$ & $9007 \pm 67$ & $-2.15 \pm 0.01$ & $13403 \pm 80$ &  $-2.22 \pm 0.01$ & $13678 \pm 81$ \\
Hallucinated            & $-5.09 \pm 0.01$ & $10458 \pm 29$ & $-5.59 \pm 0.01$ & $11686 \pm 31$ & $-2.48 \pm 0.01$ & $14219 \pm 31$ &  $-2.55 \pm 0.01$ & $14614 \pm 32$ \\
\hspace{3mm}Initial    & $-5.62 \pm 0.03$ & $13115 \pm 87$ & $-6.17 \pm 0.03$ & $14387 \pm 91$ & $-3.48 \pm 0.02$ & $16029 \pm 94$ &  $-3.61 \pm 0.02$ & $16581 \pm 96$ \\
\hspace{3mm}Subsequent & $-5.01 \pm 0.01$ & $10091 \pm 30$ & $-5.51 \pm 0.01$ & $11312 \pm 32$ & $-2.35 \pm 0.01$ & $13994 \pm 33$ &  $-2.42 \pm 0.01$ & $14370 \pm 34$ \\
\bottomrule
\end{tabular}
}
\caption{Average score and ranking values with standard error by hallucination label scored on ground truth tokens for both models under beam search and \ourdecoding.}
\label{tab:hallucination-results}
\end{table*}

\begin{table}
\centering
\resizebox{\linewidth}{!}{
\begin{tabular}{l| c | c }
\toprule
 & \multicolumn{2}{c}{Average Conditional Entropy}\\
\midrule
Token label & \transmodel & \bartmodel\\
\midrule
Non-Hallucinated        & $2.3898 \pm 0.0131$ & $3.8111 \pm 0.0292$ \\
Hallucinated            & $2.5405 \pm 0.0200$ & $3.6893 \pm 0.0209$ \\
\hspace{3mm}Initial     & $3.1147 \pm 0.0514$ & $4.1972 \pm 0.0648$ \\
\hspace{3mm}Subsequent  & $2.4490 \pm 0.0214$ & $3.7405 \pm 0.0323$ \\
\bottomrule
\end{tabular}
}
\caption{Average conditional entropy with standard error by hallucination label scored on ground truth tokens for both models under beam search.}
\label{tab:entropy-results}
\end{table}

\begin{figure}
    \centering
    \resizebox{\linewidth}{!}{
    \begin{tabular}{ p{10cm} }
    \thickhline
    \textbf{\textsc{Reference Summary:}} A drunk man who was driving his car at 119mph when he crashed into and killed an off-duty police community support officer (PCSO) has been jailed. \\
    
    \thickhline

    \textbf{\textsc{Document:}} Alwyn Pritchard, 53, was riding his motorbike when he was struck by an Audi driven by Paul Wilson, who then fled the scene, Cardiff Crown Court heard. \\ 
      
    [{\em 2 sentences with 41 words are abbreviated from here.}] \\
    
    Wilson, an experienced HGV driver, admitted drinking "a couple of pints of Peroni and two bottles of Corona" but claimed he had been driving at 70mph on the Heads of the Valleys road near Abergavenny. \\
    
     [{\em 12 sentences with 232 words are abbreviated from here.}] \\
     
    Gwent Police Chief Constable Jeff Farrar described him as "a committed, kind and conscientious community support officer".\\
    \thickhline
    \textbf{\textsc{\transmodel:}} A driver has been jailed for four years for causing the death of a man by careless driving. \\
    
    \thickhline
    \textbf{\textsc{\transmodel with \ourdecoding:}} A driver who crashed into a car which killed a couple has been jailed for seven years. \\
    
    \thickhline
    
    \textbf{\textsc{\bartmodel:}} A man has been jailed for causing the death of a police community support officer by dangerous driving in Monmouthshire. \\
    
    \thickhline
    \textbf{\textsc{\bartmodel with \ourdecoding:}} A drink-driver has been jailed for causing the death of a police community support officer in Monmouthshire.\\
    
    \thickhline
    \end{tabular}   
    }
  \caption{An abridged example from the \xsum dataset and the generated summaries under \transmodel and \bartmodel, with and without \ourdecoding decoding. 
}
\label{figure:xsum-example}
\end{figure}

\section{Implementation Details}
\label{sec:appendix-implementation}
We train all models using the Fairseq framework \cite{ott2019fairseq}. The code will be released upon acceptance. 

\paragraph{Preprocessing.}
We tokenize the data with Moses \cite{koehn2007moses}. For the \transmodel we learn and apply BPE using FastBPE \cite{sennrich2015fastbpe}, whereas for \bartmodel we follow the provided BPE preprocessing steps. \footnote{\url{https://github.com/facebookresearch/fairseq/tree/main/examples/bart}} We then binarize the resulting data using the fairseq-preprocess CLI tool from Fairseq.

\paragraph{General training and generation.}
We train on a single GPU with 4 CPU cores, each with 2048 MB memory. The average runtime for training depends on the model, but was between $24$ and $48$ hours. We stop training early if validation performance does not improve for 5 consecutive runs. We use a maximum length of 4096 tokens, and truncate longer sources. We use beam search with a beam size of 5, the same beam size is used for \ourdecoding.

\paragraph{\transmodel.}
We use the fairseq transformer model with parameters selected according to the transformer-wmt-en-de model \footnote{\url{https://github.com/facebookresearch/fairseq/tree/main/examples/translation}} We picked the parameter update frequency to be the maximum value that did not cause out-of-memory errors: $64$. We then did a grid-search over dropout in $[0.1, 0.3]$ and learning rate in  $[\num{7e-4}, \num{7e-5}]$. The optimal values were dropout of $0.3$ and learning rate of $\num{7e-5}$, with a validation loss of $6.225$.

\paragraph{\bartmodel.}
We used a pretrained BART summarization model \cite{lewis2019bart} finetuned on \xsum. 

\paragraph{Language model.} As the BPE step was different for \transmodel and \bartmodel, we trained two language models denoted by the associated summarization model name. The architecture is the fairseq transformer-lm model. The early stopping criteria was 5 runs, and maximum length was 2048 tokens. As before we picked the update-frequency to be as large as possible without taking too long, this was 32. We searched over different training sets of targets only and both source and targets.  We then did a grid-search over learning rate in  $[\num{1e-4}, \num{2.5e-4}, \num{5e-4}]$. The optimal parameters were to train on both source and targets with a learning rate of \num{1e-4} for \transmodel and \num{5e-4} for \bartmodel. The optimal validation metrics were a loss and perplexity of $5.6404$ and $49.88$ respectively for \transmodel and $4.5453$ and $23.35$ respectively for \bartmodel.

\paragraph{\ourdecoding hyperparameter search.} We select two hyperparameters $\lambda/\tau$, controlling the influence of the language model and the conditional entropy threshold which triggers PMI decoding. The goal is to perform a min max optimization, where we minimize the average log probability (scored under the PMI objective) of initial hallucinated tokens based on human evaluations of the target sentences and to maximize the \rouge-L score of generated sentences. We use the 500 example subset of \xsum with factuality annotations, that is a subset of the \xsum test set \cite{maynez2020faithfulness}. To perform the optimization we generate a heat plot with $\lambda/\tau$ on the x/y axis and the z axis is a weighted combination of \rouge score - log probability to get around a 3:1 contribution respectively to the z value. We then determine the optimal parameters to be the ones that maximize this metric. 

\begin{figure}[!ht]
    \centering
    \begin{minipage}{0.45\textwidth}
        \centering
        \includegraphics[width=0.9\textwidth]{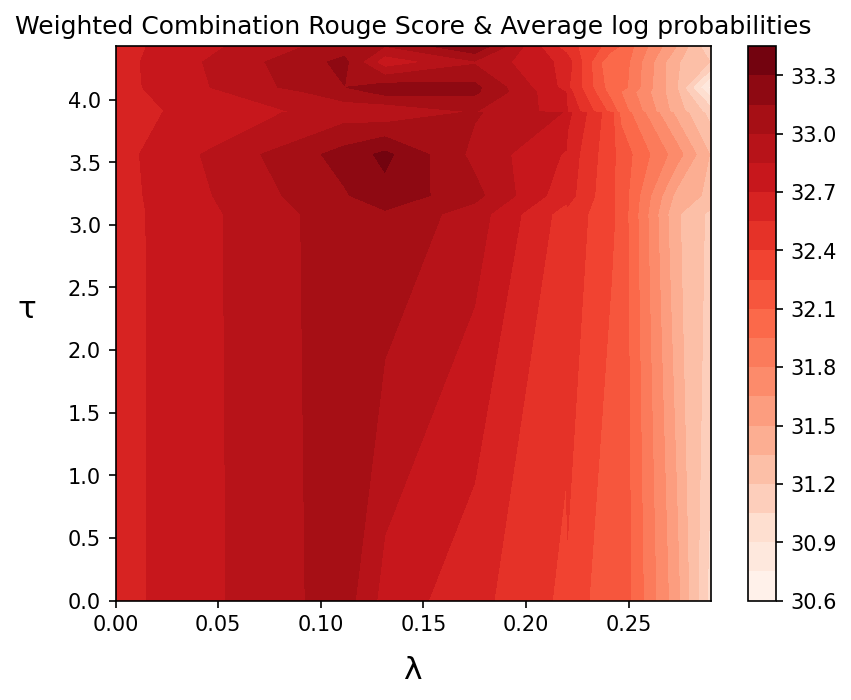}
    \end{minipage}\hfill
    \begin{minipage}{0.45\textwidth}
        \centering
        \includegraphics[width=0.9\textwidth]{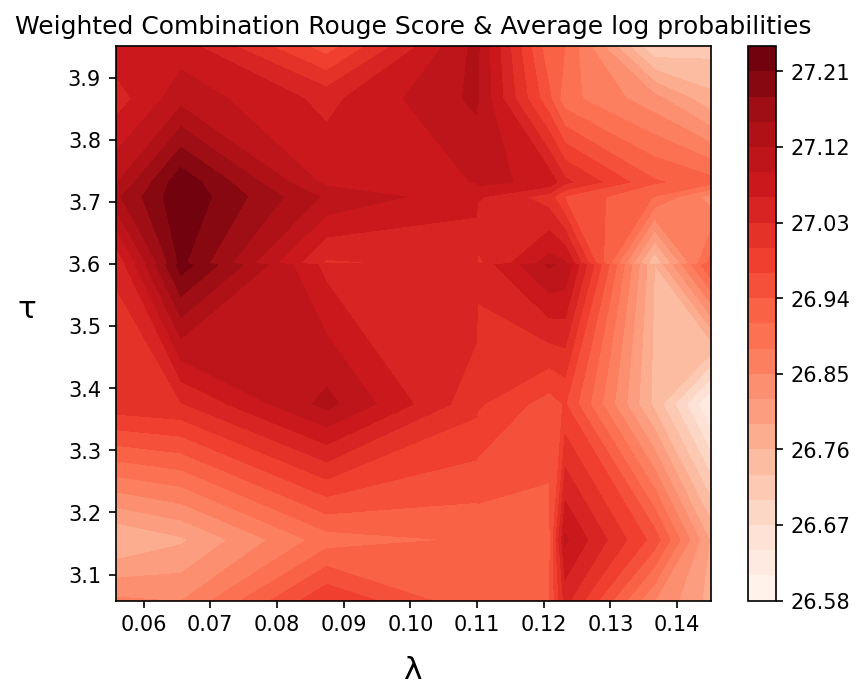}
    \end{minipage}
    \caption{Hyperparameter search for $\lambda/\tau$ for \transmodel (top) and \bartmodel (bot)}
\label{plt:hyperparam-search}
\end{figure}

There were two evaluation runs, first with $\lambda \in [\num{2e-1}, \num{2e-2}, ..., \num{2e-1}]$ and $\tau$ selected from a uniform distribution about the average conditional entropy of the initial hallucinated tokens $\pm$ the standard deviation (see \cref{tab:entropy-results} for these values). The second run, selected a smaller region the looked promising and then selected uniformly at random $10$ $\lambda/\tau$ values for a total of $100$ possible parameters pairs. The optimal values were $\lambda = \num{1.3120e-1}, \tau = 3.5618$ for \transmodel and $\lambda = \num{6.5602e-1}, \tau = 3.5987$ for \bartmodel. The plots in \cref{plt:hyperparam-search}, show the plots from this second run used to select the optimal parameters.

\paragraph{Automatic hallucination detection.} We mention in the paper that we use automatic factuality detection in order to obtain measures such as \factscore. For this we use the provided code by \citet{zhou2021detecting}. \footnote{\url{https://github.com/violet-zct/fairseq-detect-hallucination}}
\end{document}